\documentclass{ieeeaccess}
\usepackage{cite}
\usepackage{amsmath,amssymb,amsfonts}
\usepackage{algorithmic}
\usepackage{graphicx}
\usepackage{textcomp}
\usepackage{caption}
\def\BibTeX{{\rm B\kern-.05em{\sc i\kern-.025em b}\kern-.08em
    T\kern-.1667em\lower.7ex\hbox{E}\kern-.125emX}}
\begin{document}
\history{Date of publication xxxx 00, 0000, date of current version xxxx 00, 0000.}
\doi{}

\title{Deep Interactive Reinforcement Learning for Path Following of Autonomous Underwater Vehicle}
\author{\uppercase{Qilei Zhang}\authorrefmark{1},
     \uppercase{Jinying Lin}\authorrefmark{1},
\uppercase{Qixin Sha}\authorrefmark{1},
 \uppercase{Bo He}\authorrefmark{1},\IEEEmembership{Member, IEEE} and  
 \uppercase{Guangliang Li}\authorrefmark{1},\IEEEmembership{Member, IEEE}
}

\address[1]{Department of Electronic Engineering, Ocean University of China, Qingdao, China}

\markboth
{Author \headeretal: Preparation of Papers for IEEE TRANSACTIONS and JOURNALS}
{Author \headeretal: Preparation of Papers for IEEE TRANSACTIONS and JOURNALS}

\corresp{Corresponding author: Guangliang Li (e-mail: guangliangli@ouc.edu.cn).}

\begin{abstract}
Autonomous underwater vehicle (AUV) plays an increasingly important role in ocean exploration. Existing AUVs are usually not fully autonomous and generally limited to pre-planning or pre-programming tasks. Reinforcement learning (RL) and deep reinforcement learning have been introduced into the AUV design and research to improve its autonomy. However, these methods are still difficult to apply directly to the actual AUV system because of the sparse rewards and low learning efficiency. In this paper, we proposed a deep interactive reinforcement learning method for path following of AUV by combining the advantages of deep reinforcement learning and interactive RL. In addition, since the human trainer cannot provide human rewards for AUV when it is running in the ocean and AUV needs to adapt to a changing environment, we further propose a deep reinforcement learning method that learns from both human rewards and environmental rewards at the same time. We test our methods in two path following tasks---straight line and sinusoids curve following of AUV by simulating in the Gazebo platform. Our experimental results show that with our proposed deep interactive RL method, AUV can converge faster than a DQN learner from only environmental reward. Moreover, AUV learning with our deep RL from both human and environmental rewards can also achieve a similar or even better performance than that with the deep interactive RL method and can adapt to the actual environment by further learning from environmental rewards.
\end{abstract}


\titlepgskip=-15pt

\maketitle

\section{Introduction}
\label{sec:introduction}

\PARstart{I}{n} recent years, the role of autonomous underwater vehicle (AUV) in ocean exploration has become more and more important. Equipped with a series of chemical and biological sensors, AUV can conduct continuous operation without human intervention in the ocean environment. In addition, it can work independently adjusting to the changes of marine environment to complete the ocean observation task. Because of the less investment, good maneuverability and flexible control, AUV has been widely applied in many fields, such as scientific observation, resource investigation, oil and gas engineering, military applications etc.

However, today's marine applications put forward higher and higher requirements for the autonomy of AUV. Existing AUVs usually do not have good autonomy and are generally limited to pre-planning or pre-programming tasks. They work well in known and structured environments, but not in uncertain and dynamic ones. Therefore, to realize the autonomy of AUV, it is necessary for it to have strong abilities of environmental perception and understanding, online adjustment of control policies, and task planning. The path planning and following of AUV, which determines the application prospect of AUV in the marine field, can only be realized with accurate control technology, in consideration of its energy consumption, motion characteristics, speed constraints, etc. Therefore, autonomous control that can adapt to the changes of marine environment is the core technology to realize the autonomy of AUV.

PID control is the most popular traditional control methods and has been successfully applied to AUV \cite{ang2005pid}. However, the traditional control methods cannot respond and adjust to unpredictable environmental changes in real time, and cannot meet the autonomy of AUV. On the other hand, robot learning based on reinforcement learning (RL) \cite{sutton1998reinforcement} has been introduced into the AUV design and research to improve its autonomy \cite{kober2013reinforcement}. Reinforcement learning is a method for a robot controller to learn optimal control policy through interaction with the environment. The policy defines which action the controller should take when the robot is in a certain environmental state. Under the current policy, after the controller tries to select and execute an action in a certain state, it will receive a reward signal provided by the reward function defined in advance by the designer in the environment. This reward signal reflects the quality of the actions performed by the controller and is used to update the control policy. The goal of the controller is to learn a policy that maximizes the total cumulative reward.

Recently, Yu et al. \cite{yu2017deep} applied the latest deep reinforcement learning (DRL) developed by researchers at Google DeepMind \cite{mnih2015human} to AUV path following task. Deep reinforcement learning combines the advantages of deep learning (DL) \cite{schmidhuber2015deep} and reinforcement learning, and can realize the end-to-end autonomous learning and control with the raw high-dimensional environment perception information input to the behavior action. Yu et al. claimed that AUV with DRL can achieve better control effect than a PID controller in simulation experiments. However, because it is difficult to define an effective reward function and it usually provides very sparse reward signals, the robot needs a lot of time and samples to explore and test before learning an effective control policy. Therefore, the traditional reinforcement learning and deep reinforcement learning methods are still difficult to apply directly to the actual AUV system.

In order to speed up the robot learning, researchers propose interactive reinforcement learning (IRL) \cite{li2019human} based on reward shaping \cite{ng1999policy} in traditional reinforcement learning methods. Interactive reinforcement learning allows designers and even non-technical personnel to train robots by evaluating their behavior. In this way, human experience knowledge can be embedded into autonomous learning of robot to speed up its learning.

Therefore, in this paper, we combine the advantages of deep reinforcement learning and interactive RL by proposing deep interactive reinforcement learning for AUV path following tasks. In addition, since the human trainer cannot provide human rewards for AUV and AUV needs to adapt to a changing environment when it is running in the ocean, we propose a deep reinforcement learning method that learns from both human rewards and environmental rewards at the same time. We test our methods in two tasks---straight line and sinusoids curve following of AUV by simulating in Gazebo. The experimental results show that AUV with our deep interactive RL method learns much faster than with a DQN learner from environmental reward. Moreover, our deep RL learning from both human and environmental rewards can also achieve a similar or even better performance than the deep interactive RL method and adapt to the actual environment by learning from environmental rewards.

\section{Related Work}

PID controller is the most popular traditional controller for AUV and has been successfully applied in many control engineering tasks \cite{ang2005pid}. The disadvantage of PID controller is that its effect will be affected by the disturbance in the complex environment. However, due to the simple structure, the control of many underwater robots is still designed with PID controller. Although the traditional control methods such as PID controller can basically meet the control requirements of AUV, they cannot adapt to unpredictable environmental changes in real time and cope with the uncertainty of external environment.  Therefore, they cannot realize the autonomy of AUV.

Reinforcement learning (RL) \cite{sutton1998reinforcement} has been successfully applied in many robot tasks \cite{kober2013reinforcement}, and has been introduced into the autonomous control of AUV. Compared with the traditional control methods of AUV, a robot with reinforcement learning can achieve online parameter adjustment and can well cope with environmental changes and uncertainties. Even in the absence of accurate system model or high coupling system, root learning with RL can also obtain good control effect.
For example, Yuh proposed an adaptive control algorithm based on neural network \cite{yuh1994learning}. In the algorithm, a cost function is designed as reward function and the reinforcement learning method is used to realize adaptive control. Different from the traditional adaptive control method, Yuh and other simulation experiments show that AUV with reinforcement learning  can not only deal with the environmental changes and uncertainties, but also has good robustness to the unmodeled system dynamics. El Fakdi and Carreras applied the policy gradient based reinforcement learning method to the underwater robot in the submarine cable tracking task, and showed good tracking results \cite{el2008policy,carreras2013behavior}.

In addition, similar to the Actor-Critic algorithm \cite{grondman2012survey} in reinforcement learning, Cui et al. \cite{cui2017adaptive} designed two neural network models using reinforcement learning: one is used to evaluate the long-term control performance of the system; the other is used to compensate the unknown dynamic of the system, such as unknown nonlinear characteristics and interference. The simulation results of Cui et al. show that their proposed reinforcement learning algorithm based on the dual neural network converges faster than the general neural network model and PD controller, and has better stability and control effect. In addition, in order to solve the problem of high-dimensional sensory information input, Yu et al. \cite{yu2017deep} applied the deep reinforcement learning method \cite{mnih2015human} to AUV path following task, which can achieve better control effect than PID control in their simulation experiments.

However, in traditional reinforcement learning and deep reinforcement learning, the reward function is pre-defined by the agent designer before robot learning, which determines the quality of the final learned control policy and the learning speed to a large extent. To define an effective reward function is not easy, which often needs many debugging and is very time-consuming. More importantly, much experience and knowledge are difficult to be embedded in an effective reward function. An inefficient reward function means sparse reward signal and low learning efficiency. The controller needs a lot of learning samples and time to test and explore before learning an optimal policy, which seriously limits the application of the reinforcement learning method to the actual AUV control system.

Based on traditional reinforcement learning, researchers proposed interactive reinforcement learning method \cite{li2019human}, which allows people to observe and evaluate the robot's behavior, and use the evaluation as a reward signal to teach the robot how to perform tasks. A robot with interactive reinforcement learning does not need to define reward function in advance, but allows a human trainer to provide reward signal by evaluating the robot's behavior according to one's own experience and knowledge. The human reward signal reflects the quality of robot's behavior, and is used by the robot to improve its behavior. Therefore, the robot can easily use people's experience and knowledge to accelerate its learning.

Thomaz and Breazeal \cite{thomaz2008teachable} implemented a tabular \emph{Q-learning} \cite{watkins1992q} agent which can learn from both human and environmental rewards by maximizing the accumulated discounted sum of them. They also show that an agent's performance can be further improved if the human teacher was allowed to provide action advice besides human rewards. Knox and Stone \cite{knox2009interactively,knox2012learning} proposed the TAMER framework---a typical interactive reinforcement learning method. The TAMER framework trains a regression model to represent a reward function that gives a reward consistent with the human's feedback. TAMER has been tested in many simulation domains \cite{knox2012learning}, such as Cart Pole, Grid World, Tetris etc., and even in a real robot navigation task \cite{knox2013training}. The results showed that agents with TAMER can learn faster than those with traditional reinforcement learning. In order to make full use of the information in human feedback, Loftin et al. \cite{loftin2016learning} interpreted human reward as a kind of categorical feedback strategy. In addition, unlike TAMER and the work of Loftin et al., which take human reward signal as feedback to the optimal control policy expected by the trainer, Macglashan et al. \cite{macglashan2017interactive} proposed the COACH algorithm by further interpreting human reward signal as feedback to the control policy that the robot is executing, and test it in the robot navigation task. Their results show that COACH can make the robot more effective in autonomous learning.

Carrera and Ahmadzadeh et al. \cite{carrera2013towards,carrera2014learning,ahmadzadeh2013autonomous} used a imitation learning method \cite{argall2009survey} to achieve the autonomous valve rotation task for AUV. They first asked the trainer to provide a demonstration on how to operate the valve based on their knowledge, and the AUV recorded the operation and learned how to turn the valve using a dynamic primitive method. However, this only maximizes the ability of AUV to perform the task and cannot be applied to other tasks or adapt to changes in the marine environment.

\section{BACKGROUND}

\begin{figure}[htb]
 \centering
 \includegraphics[width=0.9\linewidth,height=0.5\linewidth]{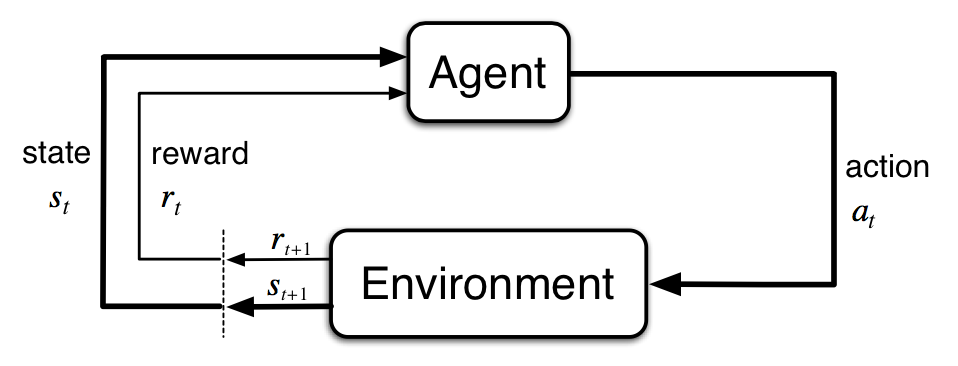}
 \caption{Illustration of learning mechanism in reinforcement learning. }
 \label{rl}
\end{figure}

As a branch of machine learning, reinforcement learning is usually modeled as an Markov decision process (MDP) \cite{sutton1998reinforcement}. MDP mainly consists of five elements: agent, environment, state, action and reward. In reinforcement learning, an agent interacts with the environment by acquiring the environment state, performing actions and obtaining rewards. The basic framework of reinforcement learning is shown in Figure \ref{rl}. Suppose that the environmental state at time $t$ is $s_t$, and the agent performs an action $a_t$ after obtaining the state $s_t$, and the environmental state is transformed from $s_t$ to $s_{t+1}$ at time $t+1$. Then the environment generates feedback reward $r_{t+1}$ to the agent in the new state $s_{t+1}$. The agent will update the learned policy with the reward signal and perform a new action $a_{t+1}$ in the new state. The agent will optimizes the policy by continually interacting with the environment until an optimal policy is learned. The agent's goal is to maximize the long-term cumulative rewards.

In reinforcement learning, a general policy $\pi$ maps states to actions and has Markov property. The probability of taking action $a$ in the current state is only related to the current state, and has nothing to do with other factors. The policy can be formulated as:
\begin{equation}
\pi(a|s)=p(a_t=a|s_t=s).
\end{equation}
Maximizing a long-term cumulative rewards means the agent needs to consider the rewards of the current time step as well as the rewards of the future time. Assume at current time step $t$ and long-term cumulative rewards can be formulated as $R=r_t+r_{t+1}+...+r_n$. However, due to the uncertainty of cumulative rewards, discounted future cumulative rewards $G_t$ is generally used in actual tasks:
\begin{equation}
G_t=R_{t+1}+\gamma R_{t+2}+\gamma^2 R_{t+3}+...=R_{t+1}+\gamma G_{t+1},
\end{equation}
where $\gamma$ is the discount factor, ranging from 0 to 1.

A value function with discounted cumulative rewards is usually used to express the degree of goodness or badness for an agent in a certain state. There are mainly two kinds of value functions: state value function and action value function. A state value function $V(s)$ represents the expectation of discounted cumulative rewards for an agent in state $s$:
\begin{equation}
V(s)=E[G_t|s_t=s],
\end{equation}
where $G_t$ represents the discount accumulated reward in the future. The action value function $Q(s,a)$ represents the reward expectation of discounted cumulative rewards for an agent performing an action $a_t$ in a certain state $s_t$ and following the  policy $\pi(s)$ thereafter:

\begin{equation}
Q(s,a)=E[G_t|s_t=s,a_t=a].
\end{equation}
The relation of action value function $Q(s,a)$ and state value function $V(s)$ can be formulated as:
\begin{equation}
V(s)=\sum_{a\in A}\pi(a|s)Q_\pi(s,a).
\end{equation}
A Bellman equation is formulated to express the relationship between the state values of state $s_t$ and state $s_{t+1}$:
\begin{equation}
V_\pi(s)=E[R_{t+1}+\gamma V_\pi(s_{t+1})|s_t=s].
\end{equation}

The objective of reinforcement learning is to find an optimal policy expressed as $\pi^*$, which can be obtained by maximizing the value function V(s) or $Q(s,a)$ under all policies:

\begin{equation}
V^*(s)=max_\pi V(s),
\end{equation}

\begin{equation}
Q^*(s,a)=max_\pi Q(s,a).
\end{equation}
Therefore, once the optimal value function is obtained, the optimal policy can be found by greedily selecting actions according to it.

By combining reinforcement learning with deep learning \cite{schmidhuber2015deep}, deep reinforcement learning (DRL) \cite{mnih2015human} fully embodies the perceived advantages of representation learning in deep learning and decision-making in reinforcement learning. In deep reinforcement learning, the problem is still defined by reinforcement learning, and the policy and value function can be represented with deep neural network and optimized based on an objective function. 

\subsection{Deep Q-network}
Mnih et al. proposed the first deep reinforcement learning algorithm --- deep Q-network (DQN) and tested in the Atari games \cite{mnih2015human}. It combines deep learning and reinforcement learning to successfully learn control policies directly from raw high-dimensional inputs. Specifically, DQN combines the neural network and Q-learning algorithm \cite{watkins1992q} to fit the Q value of each action with the deep neural network. The state value function $V_\pi(s)$ and action value function $Q_\pi(s,a)$ are approximated as $\hat{V}(s,\theta)$ and $\hat{Q}(s,a,\theta)$, where $\theta$ represents the weights. For the state value function $\hat{V}(s,\theta)$, the input of the neural network is the eigenvector of state $s$, and the output is the corresponding value function. For the action value function, the input is the feature vector of the state, and the output is the action value function with each action corresponds to an output.

In DQN, a value function represented with deep neural network is learned and optimized with state data. The DQN algorithm uses  two networks for learning: a prediction network $Q(s,a,\theta)$ and target network $Q^{'}(s,a,{\theta}^{'})$. The prediction network is used to evaluate the current state action and updated at each time step. The target network $Q^{'}(s,a,{\theta}^{'})$ is used to generate target value. The target network is directly copied from the prediction network every certain number of time steps and does not update its parameters. The objective of introducing the target network in DQN is to keep the target Q value unchanged for a period of time. In this case, the correlation between the field prediction Q value and the target Q value will be correlated to some extent, and the network instability during training can be reduced. 

Specifically, a DQN agent will select the action with the largest Q value based on the outputted Q value by the network. The experience replay mechanism \cite{lin1992self,schaul2015prioritized} is also used in DQN. It stores the experience samples $\left\{s, a, r, s^{'},ifend\right\}$(ifend flag indicates whether the task is completed) obtained from the interaction between the agent and environment at each time step into the experience replay pool. When training the network, a small batch of samples is randomly selected from the experience replay pool and the current target Q value y is computed as:

\begin{equation}
y_i=\left\{
             \begin{array}{ll}
             R_i, & ifend=true\\
             R_i+\gamma max_a\hat{Q^{'}}(s^{'},a^{'},{\theta}^{'}), & ifend=false
             \end{array}.
             \right.
\end{equation}
This helps remove the correlation and dependence between samples and makes the network more convergent.
With the mean square error loss function
\begin{equation}L=\dfrac{1}{N}\sum_{i=1}^N(y_i-\hat{Q}(s,a,\theta))^2,
\end{equation}
all parameters $\theta$ of Q network will be updated through gradient back propagation\cite{riedmiller1993direct}.

\subsection{Interactive reinforcement learning}

\begin{figure}[htb]
 \centering
 \includegraphics[width=0.85\linewidth,height=0.65\linewidth]{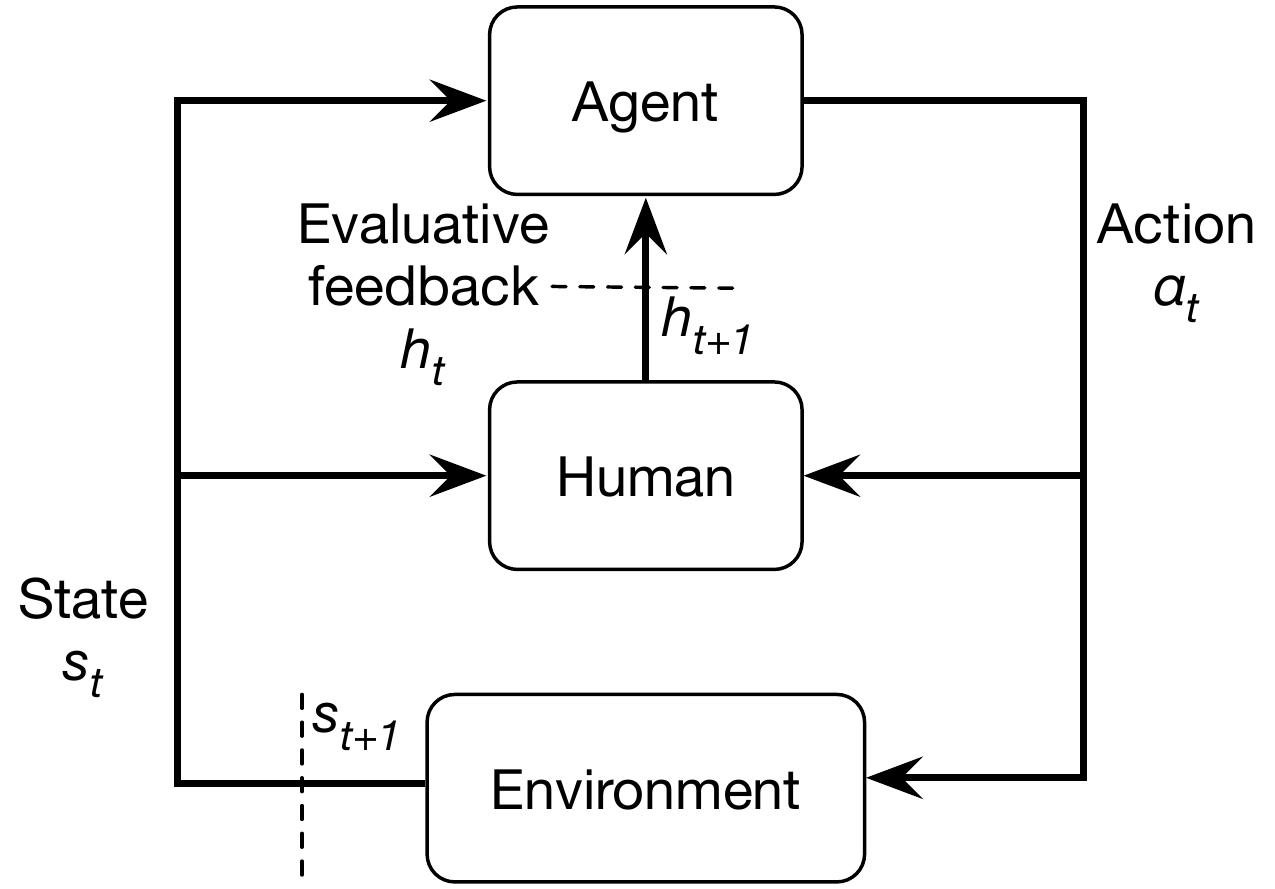}
 \caption{Illustration of interactive reinforcement learning with human feedback. }
 \label{irl}
\end{figure}

In reinforcement learning and deep reinforcement learning, the reward function is pre-defined by the controller designer before agent learning. The reward function determines the quality of the final control policy and the learning speed to a large extent. However, it is not easy to define an effective reward function since it often needs many debugging, and much experience and knowledge are difficult to be embedded in an effective reward function. An inefficient reward function means that the controller needs a lot of learning samples and time to test and explore, which seriously limits the application of the traditional reinforcement learning method to the actual robot system.

Therefore, on the basis of traditional reinforcement learning, researchers put forward interactive reinforcement learning \cite{li2019human,thomaz2008teachable,knox2009interactively,knox2012learning,loftin2016learning,li2016using,li2016socially,macglashan2017interactive,li2018social}. In interactive reinforcement learning, an agent learns in an MDP without reward function. It allows people to observe and evaluate the agent's behavior, and give a reward signal based on their judgement of agent's action selection. The stand or fall of reward signal reflects the quality of agent's behavior. The agent uses the reward signal delivered by the human trainer to improve its behavior. The learning mechanism of interactive reinforcement learning is shown in Figure \ref{irl}.

\section{Approach}

\begin{figure*}[htb]
 \centering
 \includegraphics[width=0.7\linewidth,height=0.5\linewidth]{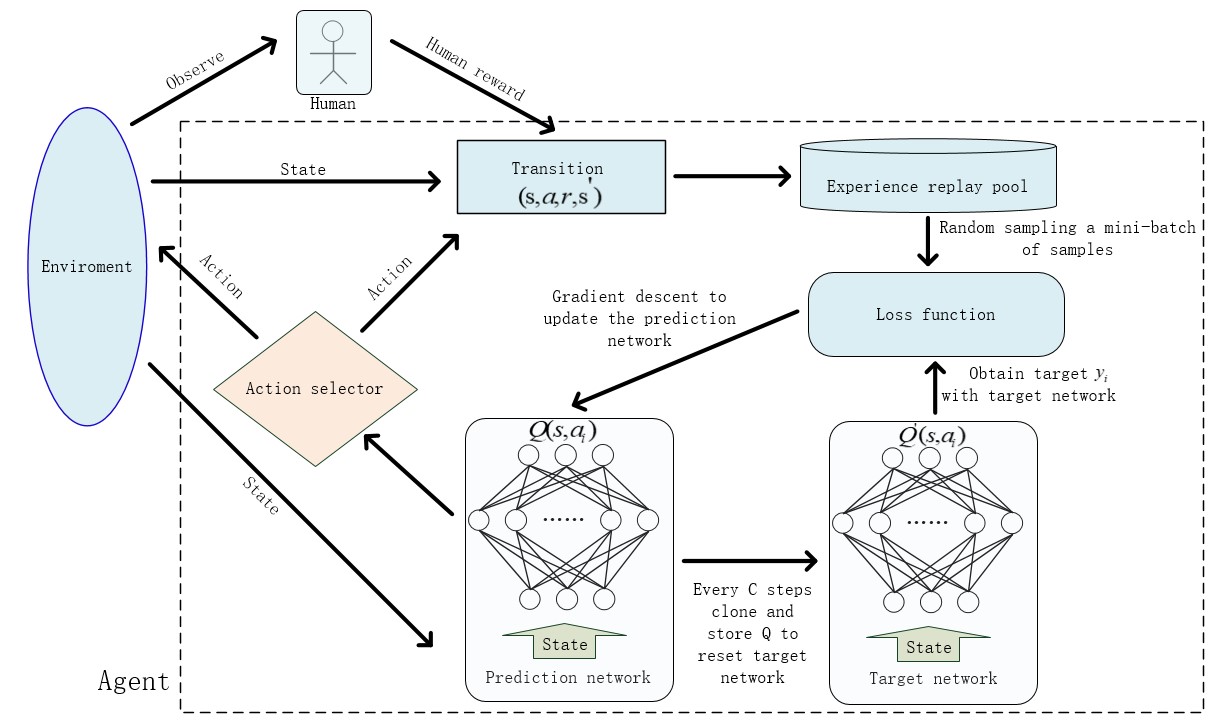}
 \caption{Learning mechanism of deep interactive reinforcement learning algorithm.}
 \label{dirl}
\end{figure*}

In this paper, we propose deep interactive reinforcement learning to complete the underwater vehicle path following task. In our proposed method, we use the DQN algorithm. However, the robot learns from human reward instead of pre-defined environment reward as in the original DQN. In our proposed method, the trainer provides rewards $R_h$ observing the running state of AUV, and evaluating the actions selected by AUV in the current state according to her knowledge. The learning mechanism of DQN is shown in Figure \ref{dirl}.

In addition, since the human trainer cannot provide rewards to train AUV all the time and AUV also needs to adapt to the changing ocean environment in the task, we propose to allow our proposed method to learn from both human reward and environment reward at the same time.

\begin{figure}[htb]
 \centering
 \includegraphics[width=1\linewidth,height=0.75\linewidth]{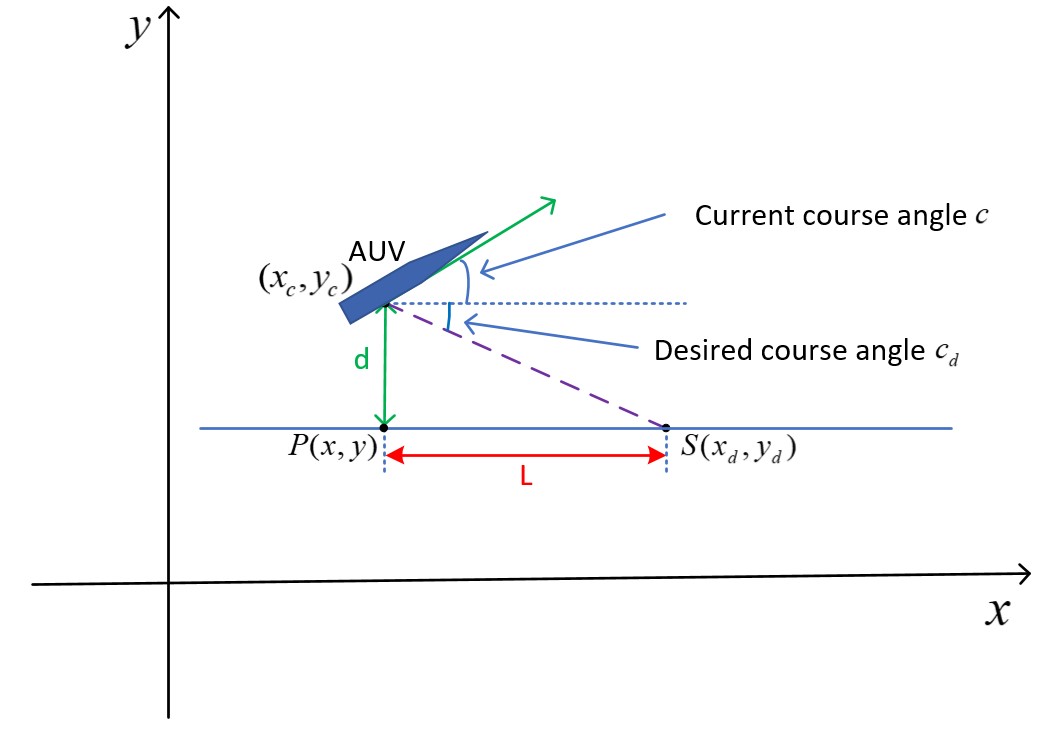}
 \caption{The desired course angle in the straight line following task, where $L$ represents the current target path, and $d$ represents the distance from AUV to the target line, $S$ represents the current target point.}
 \label{ecl}
\end{figure}

\begin{figure}[htb]
 \centering
 \includegraphics[width=1\linewidth,height=0.75\linewidth]{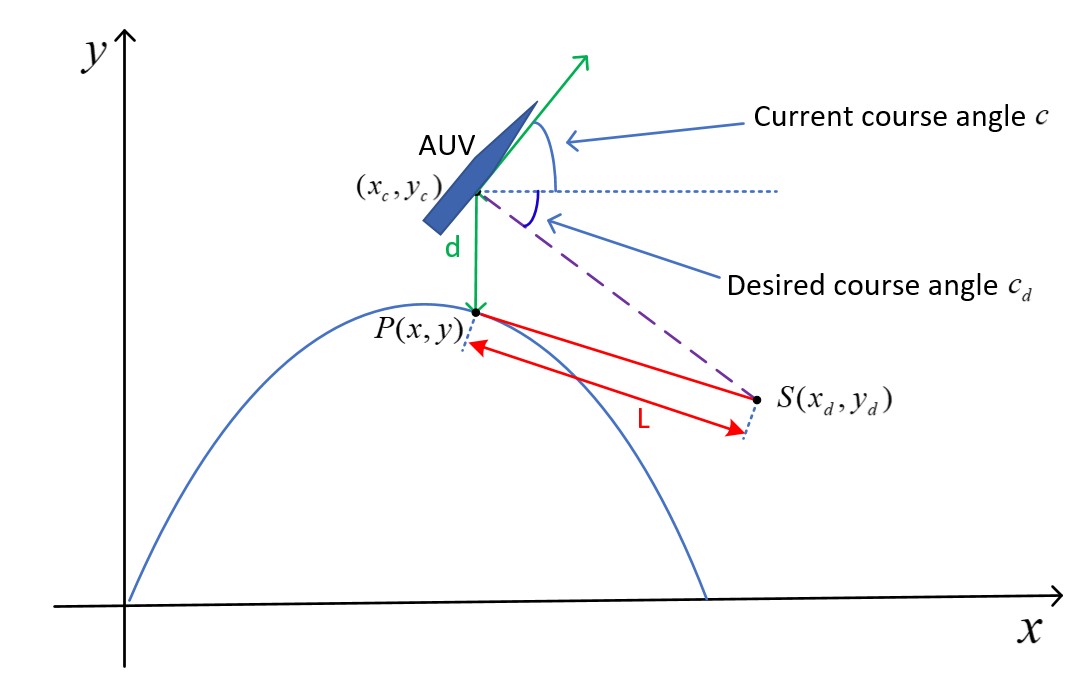}
 \caption{The desired course angle in the sinusoids curve following task, where $L$ represents the current target path, $d$ represents the distance from AUV to the target curve, $S$ represents the current target point.}
 \label{ecc}
\end{figure}

We test our methods in two path following tasks: straight line and sinusoids curve following. In both tasks, the input to the behavior value function network of DQN is the state of AUV. In the straight line following task, the input state $s$ is represented by the current course of AUV and the shortest distance from AUV to the target path: $S=\left\{d\quad c\right\}$, where $d$ represents the distance from the position of AUV to P(x,y)---the intersection point of the perpendicular line to the horizontal axis at the position of AUV and the target line, $c$ represents the current course angle of AUV, as shown in Figure \ref{ecl}. In the sinusoids curve following task, the slope of the tangent line at the intersection point  P(x,y) and desired course angle are added as additional features to represent the state: $S=\left\{d\quad c\quad k\quad c_d\right\}$, where $d$ represents the distance from the position of AUV to the intersection point P(x,y), $c$ represents current course angle, $k$ represents the slope of the tangent line at the intersection point P(x,y), $c_d$ represents desired course angle, as shown in Figure \ref{ecc}. The action space is the rudder value of AUV, and different actions correspond to different rudder angle values.

The environment reward function is defined as the difference between the current course angle and the desired course angle of AUV, considering the distance from AUV to the target line/curve. The way we calculate the desired course angle is similar to the line of sight (LOS) algorithm \cite{fossen2003line}. For example, in the straight line following task, we choose the current target path $L$ with a fixed value from the intersection point $P(x,y)$ along the target line. Then the destination point $S(x_d,y_d)$ is decided since the length of the current path $L$ is fixed. The desired course angle is computed as the angle between the desired course from the current position of AUV to the destination point $S(x_d,y_d)$ and the horizontal axis, as shown in Figure \ref{ecl}.

In the sinusoids curve following task, the computation of the desired course angle is a bit different from the straight line following task. Specifically, we choose current target path $L$ with a fixed length along the tangent line at the intersection point $P(x,y)$. Then the desired course angle is computed as the angle between the desired course from the current position of AUV to the current target point $S(x_d,y_d)$ and the horizontal axis, as shown in Figure \ref{ecc}.

If the distance between the AUV and the following line or curve is directly added to the reward function, it may lead to a large fluctuation of the reward value. Therefore, we introduce an exponential transformation to define the reward function as: 

\begin{equation}
R=-0.9\cdot\left|c_d-c\right|+0.1\cdot2^{2-\frac{d}{10}},
 \label{er}
\end{equation}
where $c$ represents the actual current course angle of AUV, $d$ represents the actual distance from AUV to the target line/curve, $c_d$ indicates the desired course angle. In this case, when the action selected by AUV reduces the difference between the current course and the desired course or the distance between AUV and the target path is smaller, the received reward R will be larger, otherwise R will be smaller. During the learning process, the experience received feedback is stored as a sample in the experience replay pool. A small batch of samples in the experience replay pool is randomly selected to update the parameters of Q network as the DQN algorithm.

\section{Experiments}

\begin{figure}[htb]
 \centering
 \includegraphics[width=0.9\linewidth,height=0.7\linewidth]{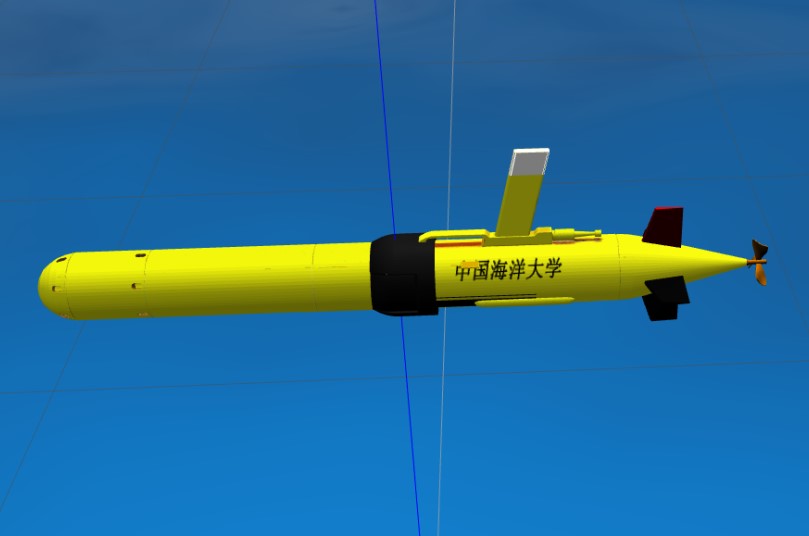}
 \caption{The autonomous underwater vehicle simulator in the Gazebo robotic platform used in our experiment.}
 \label{auv}
\end{figure}

To verify the effectiveness of our proposed method, we conducted experiments with an extension to the open-source robot simulator Gazebo in underwater scenarios. The Autonomous Underwater Vehicle (AUV) Simulator \cite{Manhaes_2016} was used in our experiment. However, we modified it to make fit with the actual AUV system in our lab, as shown in Figure \ref{auv}. The AUV simulator uses the robot operating system (ROS) to communicate with the underwater environment. It can also simulate multiple underwater robots and intervention tasks using robotic manipulators.

\begin{figure}[htb]
 \centering
 \includegraphics[width=0.9\linewidth,height=0.7\linewidth]{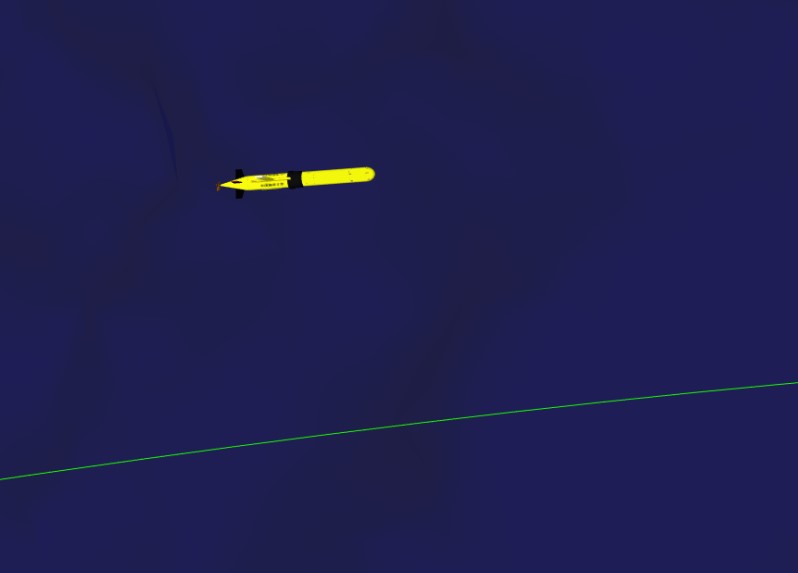}
 \caption{Human-machine interaction interface for displaying of the real-time attitude of AUV. Note that the solid green line represents target path. }
 \label{display}
 \vspace{-4mm}
\end{figure}

In our proposed deep interactive reinforcement learning method, the human trainer needs to evaluate the state and action of AUV. It is not easy to observe the exact action of AUV in the simulated environment. Therefore,  we developed a human-machine interaction interface for the human trainer to observe the attitude of AUV in real time using the Rviz display tool, as shown in Figure \ref{display}. Rviz is a built-in graphical tool of ROS, which makes it very convenient for users to develop and debug ROS through the graphical interface.

In our experiments, the AUV simulator selects and executes an action $a$ based on its current state $s$ and learned policy. The human trainer observes the action selected by the AUV simulator in the current state from the visualized interface and evaluates its quality according to her knowledge and experience. The evaluation will be taken as human reward and delivered to the AUV simulator via the developed human-machine interface. The AUV simulator will use it to update the DQN network parameters to improve its control policy.

We trained three agents in our experiments: the DQNH agent, a DQN agent learning from only human provided reward as we proposed; the DQNHE agent, a DQN agent learning from both environment reward and human provided reward; the DQNE agent, a baseline agent where a DQN agent learns from only environment reward. For both DQNHE and DQNE agents, the environment reward is provided via Equation \ref{er}. For the DQNH and DQNHE agents to learn from human reward, the human trainer will give a reward value +0.8 or +0.5 when she thinks the AUV's movement is good through the developed interface. When she thinks the AUV's action is bad, a reward value -0.8 or -0.5 will be given. Specific reward value can be selected by the trainer according to her experience, as below:

\begin{equation}
R_h=\left\{
             \begin{array}{lr}
             +0.8\quad  or +0.5, & good\quad action  \\
             -0.8\quad  or -0.5, & bad\quad action.
             \end{array}
\right.
\end{equation}
When the DQNHE agent learning, the human reward $R_h$ is added to the environment reward $R$ as the final reward value.

\section{Experimental Results and Discussion}

In this section, we present our experimental results tested in two path following tasks: straight line following and sinusoids curve following. We compare our proposed DQN agent learning from solely human reward, termed DQNH, DQN agent learning from both human reward and environmental reward, termed DQNHE, to the original DQN agent learning from solely environment reward, termed DQNE.

\subsection{Straight Line Following}
\begin{figure}[htb]
 \centering
 \includegraphics[width=1\linewidth,height=0.75\linewidth]{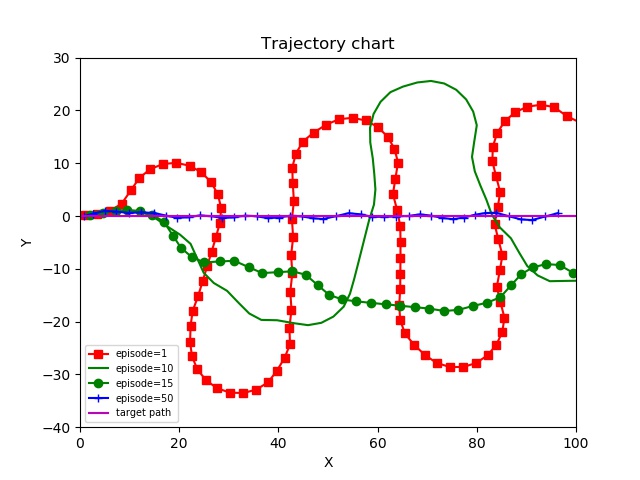}
 \caption{Trajectories of DQNE agent learning in the straight line following task. Note that X is the coordinate along the horizontal axis, Y is the coordinate along the vertical axis.}
 \label{dqne}
\end{figure}

In this experiment, we trained the DQNH, DQNHE and DQNE agents in the straight line following task. Figure \ref{dqne} shows the trajectories of the DQNE agent learning at different learning episodes in the straight line task. From Figure \ref{dqne} we can see that, at Episode 1, the trajectory of the agent fluctuate dramatically along the target line at both sides. At Episode 10, the fluctuation of the learning trajectory was reduced to some extent and at Episode 15, the agent can already follow a line but still far away from the target one. Until the 50th episode, the trajectory of the DQNE agent is quite close to the target line.

\begin{figure}[htb]
 \centering
 \includegraphics[width=1\linewidth,height=0.75\linewidth]{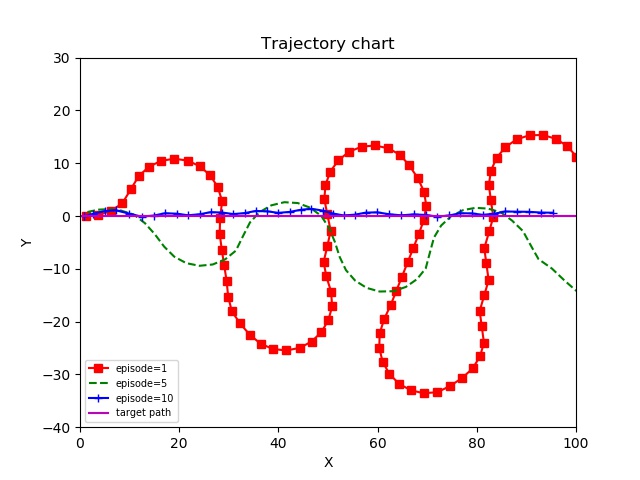}
 \caption{Trajectories of the DQNH agent learning in the straight line following task. Note that X is the coordinate along the horizontal axis, Y is the coordinate along the vertical axis.}
 \label{dqnh}
\end{figure}

Figure \ref{dqnh} shows the trajectories of the DQNH agent learning at different learning episodes in the straight line following task. From Figure \ref{dqnh} we can see that, at Episode 1, the trajectory of the DQNH agent is similar to the one of the DQNE agent and fluctuate along the target line at both sides. At Episode 5, the fluctuation of the DQNH agent learning trajectory was reduced to a large extent. At Episode 10, the trajectory of the DQNH agent is already quite close to the target line, achieving a similar performance to the DQNE agent at the 50th episode as shown in Figure \ref{dqne}. By comparing the learning trajectories of DQNH agent in Figure \ref{dqnh} to those of DQNE agent in Figure \ref{dqne}, we found that a DQN agent learning from solely human reward converges much faster than learning from solely environment reward.

\begin{figure}[htb]
 \centering
 \includegraphics[width=1\linewidth,height=0.7\linewidth]{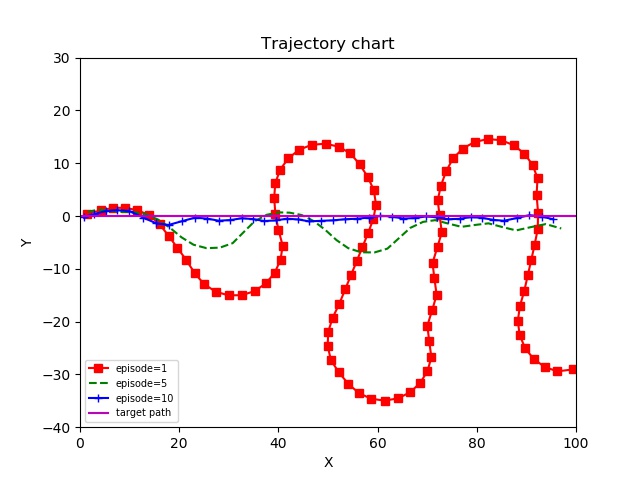}
 \caption{Trajectories of the DQNHE agent learning in the straight line following task. Note that X is the coordinate along the horizontal axis, Y is the coordinate along the vertical axis.}
 \label{dqnhe}
\end{figure}

Figure \ref{dqnhe} shows the trajectories of the DQNHE agent learning at different learning episodes in the straight line following task. From Figure \ref{dqnhe} we can see that, at Episode 1, the trajectory of the DQNHE agent is similar to the ones of the DQNE agent and DQNH agent. However, at Episode 5, the fluctuation of the DQNHE agent learning trajectory was dramatically reduced and even much better than that of the DQNH agent. At Episode 10, the trajectory of the DQNHE agent is also close to the target line, achieving a similar performance to the DQNH agent. By comparing the learning trajectories of DQNHE agent in Figure \ref{dqnhe} to those of DQNH agent in Figure \ref{dqnh}, we found that a DQN agent learning from both human reward and environment reward can further improve its convergence speed, compared to learning from solely human reward.

\begin{figure}[htb]
 \centering
 \includegraphics[width=1\linewidth,height=0.7\linewidth]{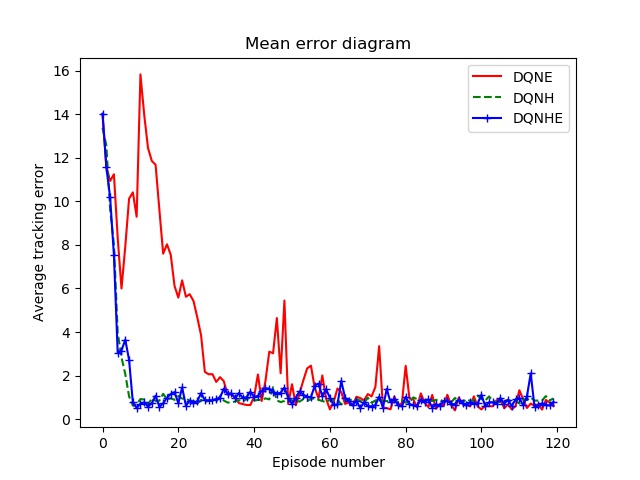}
 \caption{Tracking error of the DQNE, DQNH, DQNHE agents along the learning process in the straight line following task (averaged over data collected in two trials).}
 \label{linetrackError}
\end{figure}

We also analyzed the tracking error of the DQNE, DQNH, DQNHE agents along the learning process in the straight line following task, as shown in Figure \ref{linetrackError}. From Figure \ref{linetrackError} we can see that, when the DQN agent learning from solely human reward or both from human reward and environment reward, the tracking error was dramatically reduced to a minimum in about 10 episodes, while the DQNE agent learning from solely environment reward achieves a similar performance until around the 40th episode. 

\begin{figure}[htb]
 \centering
 \includegraphics[width=1\linewidth,height=0.7\linewidth]{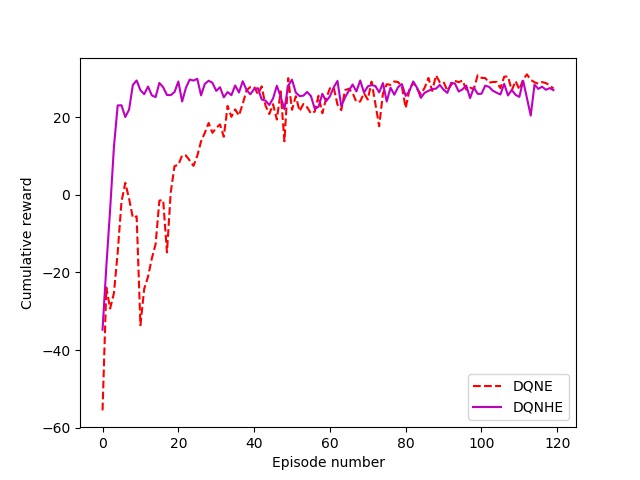}
 \caption{Cumulative environment rewards obtained by the DQNE and DQNHE agents in the straight line following task (averaged  over data collected in two trials).} 
 \label{rewardline}
\end{figure}

In addition, the cumulative environment rewards obtained by the DQNE and DQNHE agents were analyzed and illustrated in Figure \ref{rewardline}. Figure \ref{rewardline} indicates that the cumulative rewards obtained by DQNHE agent quickly reach the peak in around 10 episodes for the first time, while it takes about 40 episodes for the DQNE agent to reach a similar level. After that, both agents converged. These results suggest that the additional human reward helps the DQN agent converge faster.

\subsection{Sinusoids Curve Following}

In this experiment, we trained the DQNH, DQNHE and DQNE agents in the sinusoids curve following task, which is a much complex one than the straight line following task.

\begin{figure}[htb]
 \centering
 \includegraphics[width=1\linewidth,height=0.75\linewidth]{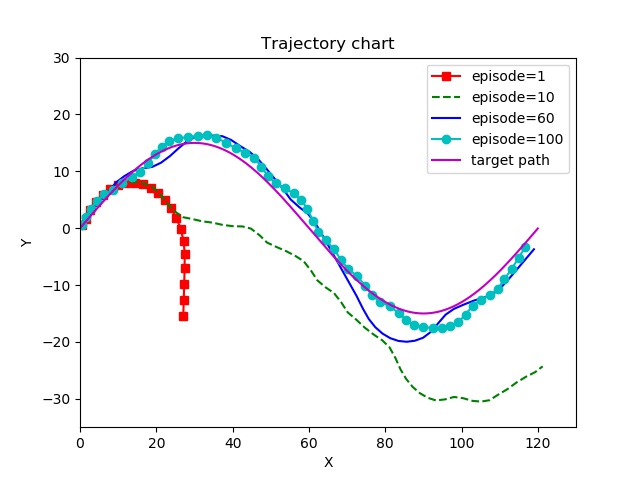}
 \caption{Trajectories of DQNE agent learning in the sinusoids curve following task. Note that X is the coordinate along the horizontal axis, Y is the coordinate along the vertical axis.} 
 \label{dqnec}
\end{figure}

Figure \ref{dqnec} shows the trajectories of the DQNE agent learning at different learning episodes in the sinusoids curve following task. From Figure \ref{dqnec} we can see that, at Episode 1 and 10, the DQNE agent cannot finish the task at all. Until the 60th episode, the fluctuation was reduced and the agent can already follow the sinusoids curve with the trajectory quite close to the target curve. At Episode 100, the agent can almost exactly follow the sinusoids curve.

\begin{figure}[htb]
 \centering
 \includegraphics[width=1\linewidth,height=0.75\linewidth]{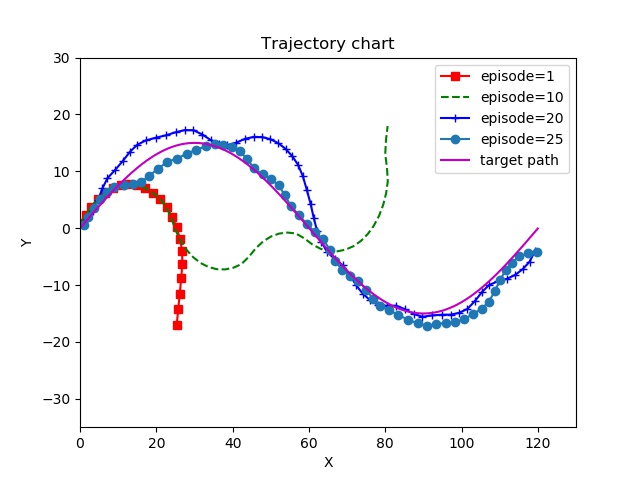}
 \caption{Trajectories of DQNH agent learning in the sinusoids curve following task. Note that X is the coordinate along the horizontal axis, Y is the coordinate along the vertical axis.}
 \label{dqnhc}
\end{figure}

Figure \ref{dqnhc} shows the trajectories of the DQNH agent learning at different learning episodes in the sinusoids curve following task. From Figure \ref{dqnhc} we can see that, at Episode 1 and 10, the DQNH agent cannot finish the task either. At 20th episode, the DQNH agent can already follow the sinusoids curve with the trajectory quite close to the target curve, achieving a similar performance to the DQNE agent at the 60th episode. At Episode 25, the AUV with DQNH can almost exactly follow the sinusoids curve, achieving a similar performance to the DQNE agent at the 100th episode.

\begin{figure}[htb]
 \centering
 \includegraphics[width=1\linewidth,height=0.7\linewidth]{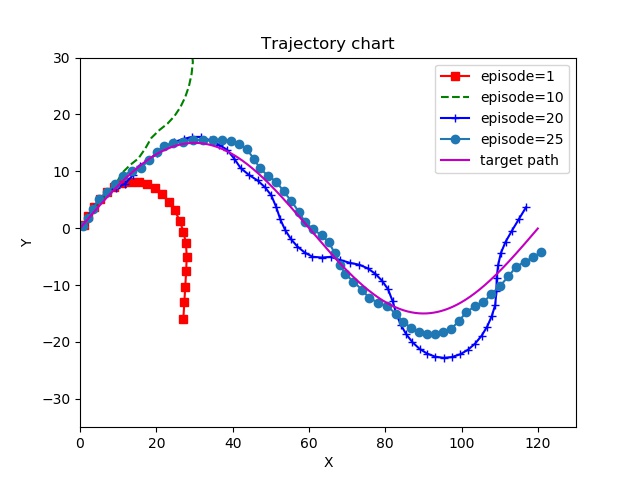}
 \caption{Trajectories of DQNHE agent learning in the sinusoids curve following task. Note that X is the coordinate along the horizontal axis, Y is the coordinate along the vertical axis.}
 \label{dqnhec}
\end{figure}

Figure \ref{dqnhec} shows the trajectories of the DQNHE agent learning at different learning episodes in the sinusoids curve following task. From Figure \ref{dqnhec} we can see that, at Episode 1 and 10, the DQNHE agent cannot finish the task either. At the 20th episode, the DQNHE agent can also follow the sinusoids curve with the trajectory quite close to the target curve. At Episode 25, the DQNHE agent can achieve a similar performance to that of the DQNE agent at Episode 100.

\begin{figure}[htb]
 \centering
 \includegraphics[width=1\linewidth,height=0.7\linewidth]{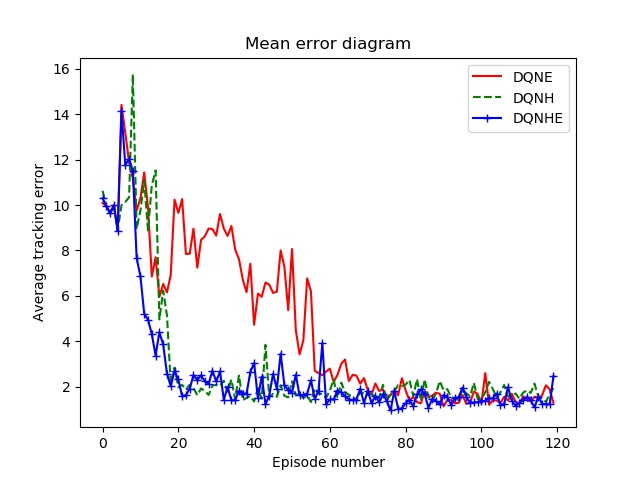}
 \caption{Tracking error of the DQNE, DQNH, DQNHE agents along the learning process in the sinusoids curve tracking task (averaged over data collected in two trials)}
 \label{trackErrorC}
\end{figure}

The tracking errors of the DQNE, DQNH, DQNHE agents along the learning process in the sinusoids curve following task were shown in Figure \ref{trackErrorC}. From Figure \ref{trackErrorC} we can see that, when the DQN agent learning from solely human reward, the tracking error was dramatically reduced to the lowest level in about 20 episodes, while the DQNE agent learning from solely environment reward achieves a similar performance until around the 70th episode. When learning both from human reward and environment reward, the tracking error is reduced the fastest.

\begin{figure}[htb]
 \vspace{-4mm}
 \centering
 \includegraphics[width=1\linewidth,height=0.7\linewidth]{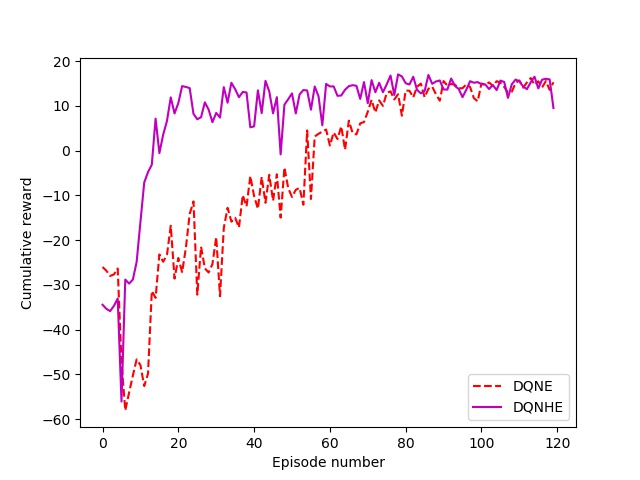}
 \caption{Cumulative environment rewards obtained of the DQNE and DQNHE agents in the sinusoids curve following task (averaged over data collected in two trials).}
 \label{rewardc}
 \vspace{-4mm}
\end{figure}

In addition, the cumulative environment rewards obtained by the DQNE and DQNHE agents were analyzed and illustrated in Figure \ref{rewardc}. Figure \ref{rewardc} indicates that the cumulative rewards obtained by DQNHE agent also quickly reach the peak in around 20 episodes for the first time, while it takes about 80 episodes for the DQNE agent to reach a similar level. After that, both agents converged.

In summary, our results in both the straight line and sinusoids curve following tasks suggest that our proposed deep interactive reinforcement learning method can facilitate a DQN agent to converge much faster than learning solely from environment reward. Moreover, our proposed DQNHE agent learning from both human reward and environment reward can combine the advantages of our proposed deep interactive reinforcement learning from human reward and learning from environment reward. The DQNHE method facilitate AUV to reach a similar performance to the one learning from solely environment reward with the speed same to or even better than the one learning from solely human reward. This allows AUV to keep at the peak performance and adapt to the environment even when the human reward is not available in the ocean.

\section{conclusion}

In this paper, we proposed a deep interactive reinforcement learning method for path following of AUV by combining the advantages of deep reinforcement learning and interactive RL. In addition, since the human trainer cannot provide human rewards for AUV all the time and AUV needs to adapt to a changing environment when it is running in the ocean, we propose a deep reinforcement learning method that learns from both human rewards and environmental rewards at the same time. We test our methods in two tasks---straight line and sinusoids curve following of AUV by simulating in the Gazebo platform. Our experimental results show that with our propose deep interactive RL method AUV can learn much faster than that a DQN learner from only environmental reward. Moreover, our deep RL learning from both human and environmental rewards can also achieve a similar or even better performance than the deep interactive RL method and adapt to the actual environment by learning from environmental rewards.

In the future, we would like to extend and test our methods with actual AUV system in ocean observation task.

\bibliographystyle{ieeetr}
\bibliography{access}

\begin{IEEEbiography}[{\includegraphics[width=1in,height=1.25in,clip,keepaspectratio]{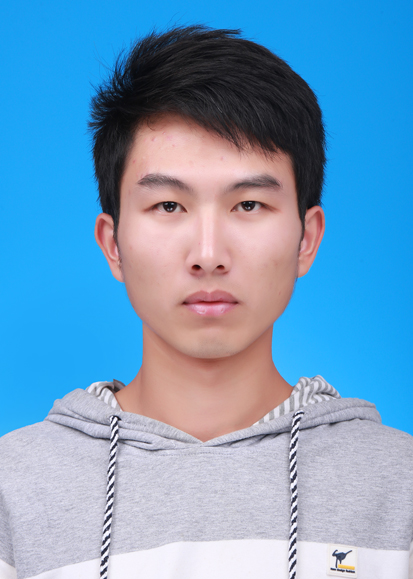}}]{Qilei Zhang} received the Bachelor's degree in electronic information science and technology from the School of Information Science and Engineering, Shandong Agricultural University, Taian, China, in 2018. He is currently a Postgraduate Student at School of Information Science and Engineering, Ocean University of China, Qingdao, China. His current research interests include reinforcement learning, human agent/robot interaction, and robotics.
\end{IEEEbiography}

\begin{IEEEbiography}[{\includegraphics[width=1in,height=1.25in,clip,keepaspectratio]{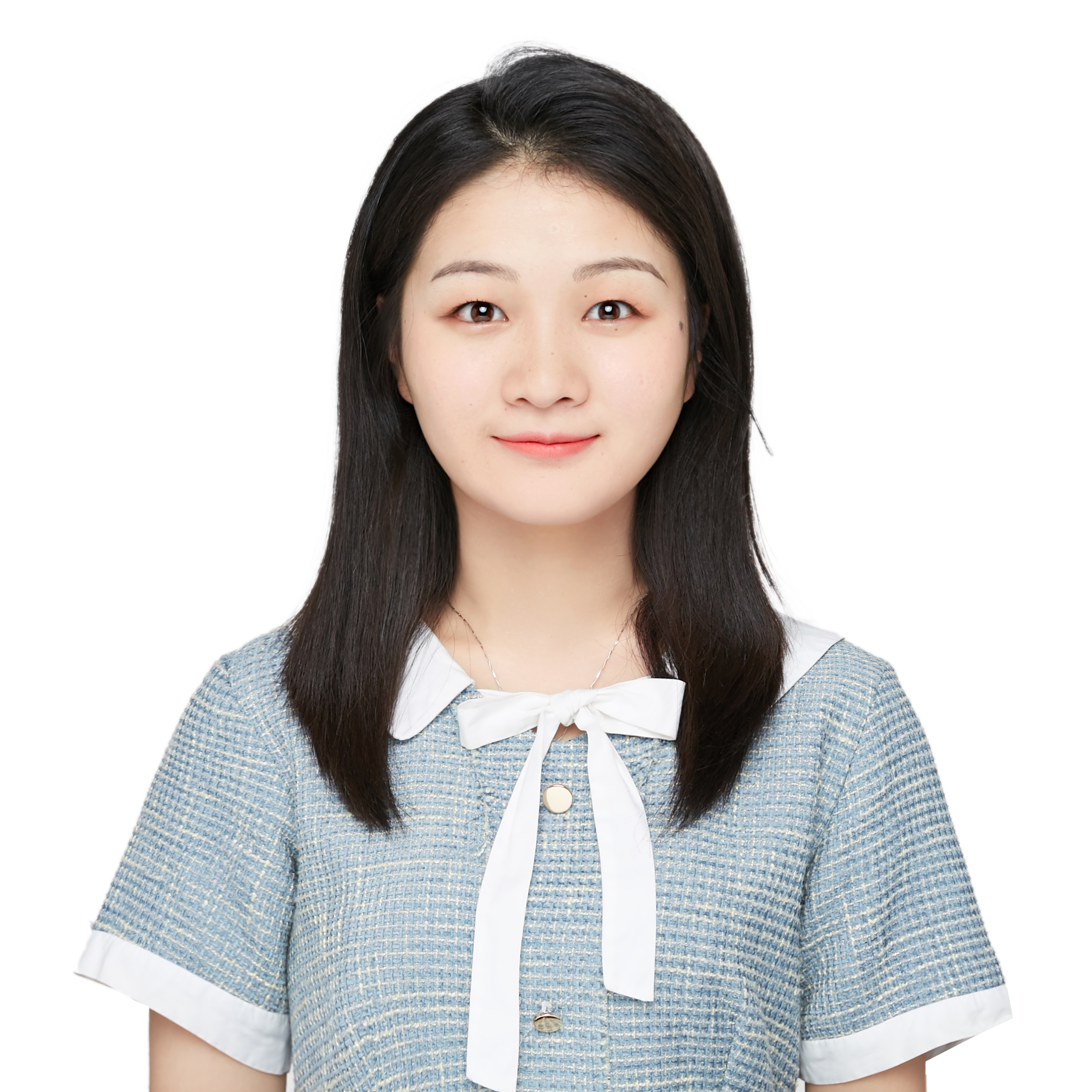}}]{Jinying Lin} received the Bachelor's degree in communication engineering from the School of Communication and Electronic Engineering, Qingdao University of Technology, Qingdao, China, in 2018. She is currently a Postgraduate Student at School of Information Science and Engineering, Ocean University of China, Qingdao, China. Her current research interests include reinforcement learning, human agent/robot interaction, and robotics.
\end{IEEEbiography}

\begin{IEEEbiography}[{\includegraphics[width=1in,height=1.25in,clip,keepaspectratio]{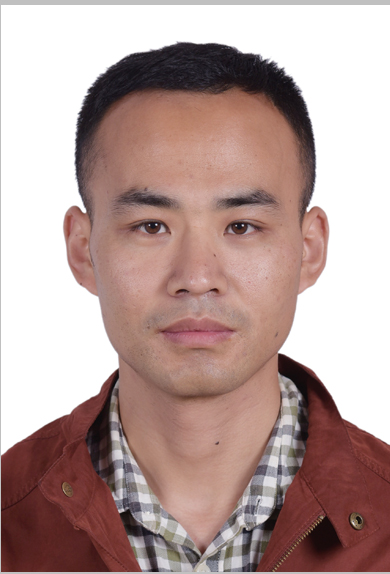}}]{Qixin Sha} received the B.S. degree in communication engineering from Ocean University of China in 2007, and the M.S. degree in communication and information systems from Ocean University of China in 2010. He worked at Qingdao Bailing Technology Co., Ltd. and Alcatel-Lucent Qingdao R \& D Center from 2010 to 2014 as a software engineer. He is currently working as an experimenter in the Department of Electronic Engineering, Ocean University of China. His research interests include the design and development of architecture, decision-making and software system in underwater vehicle.
\end{IEEEbiography}

\begin{IEEEbiography}[{\includegraphics[width=1in,height=1.25in,clip,keepaspectratio]{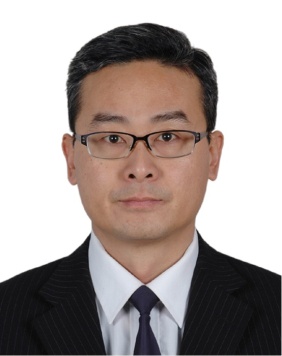}}]{Bo He} (M'18) received the Ph.D. degree in control theory and control engineering from Harbin Institute of Technology, Harbin, China, in 1999.

 He was a Researcher with Nanyang Technological University, Singapore, from 2000 to 2002. He is currently a Full Professor with Ocean University of China, Qingdao, China. His research interests include SLAM, machine learning, and robotics.
\end{IEEEbiography}

\begin{IEEEbiography}[{\includegraphics[width=1in,height=1.25in,clip,keepaspectratio]{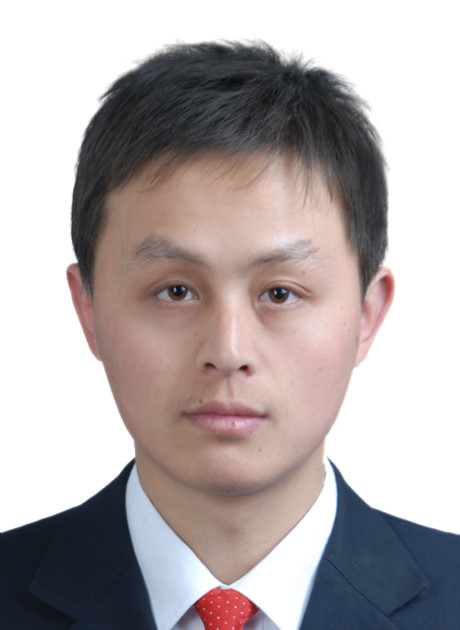}}]{Guangliang Li} (M'14) received the Bachelor's degree in automation and M.Sc. degree in control theory and control engineering from the School of Control Science and Engineering, Shandong University, Jinan, China, in 2008 and 2011, respectively, and the Ph.D. degree in computer science from the University of Amsterdam, Amsterdam, The Netherlands, in 2016.

 He was a Visiting Researcher with Delft University of Technology, Delft, The Netherlands, and a Research Intern with the Honda Research Institute Japan, Co., Ltd., Wako, Japan. He is currently a Lecturer with Ocean University of China, Qingdao, China. His research interests include reinforcement learning, human agent/robot interaction, and robotics.
\end{IEEEbiography}

\EOD
\end{document}